\documentclass[conference]{IEEEtran}
\IEEEoverridecommandlockouts
\IEEEpubid{\makebox[\columnwidth]{978-1-7281-6215-7/20/\$31.00 ~
\copyright2020 IEEE \hfill} \hspace{\columnsep}\makebox[\columnwidth]{ }}
\makeatletter
\usepackage[inline]{enumitem}
\def\BState{\State\hskip-\ALG@thistlm}
\makeatother

\newcommand{\SLR}{\mbox{\emph{SLR}}}
\newcommand{\WBSLR}{\mbox{\emph{WB-SLR}}}

\usepackage{multirow}
\usepackage{comment}
\usepackage{xcolor}
\usepackage{color}
\usepackage[ruled,vlined]{algorithm2e}
\usepackage[colorlinks=true,
            linkcolor=red,
            urlcolor=blue,
            citecolor=blue]{hyperref}
\usepackage{cite}
\usepackage{amsmath,amssymb,amsfonts}
\usepackage{algorithmic}
\usepackage{graphicx}
\usepackage{textcomp}
\usepackage{xcolor}
\def\BibTeX{{\rm B\kern-.05em{\sc i\kern-.025em b}\kern-.08em
    T\kern-.1667em\lower.7ex\hbox{E}\kern-.125emX}}
\begin{document}

\title{Sparse Longitudinal Representations of Electronic Health Record Data for the Early Detection of Chronic Kidney Disease in Diabetic Patients}

\author{\IEEEauthorblockN{Jinghe Zhang\IEEEauthorrefmark{1},
 Kamran Kowsari\IEEEauthorrefmark{1}\IEEEauthorrefmark{2},
 Mehdi Boukhechba\IEEEauthorrefmark{1}, 
 James Harrison\IEEEauthorrefmark{3},
 Jennifer Lobo\IEEEauthorrefmark{3} and
 Laura Barnes\IEEEauthorrefmark{1}}

\IEEEauthorblockA{\IEEEauthorrefmark{1} Department of System and Information Engineering,
University of Virginia,
Charlottesville, VA, USA}

\IEEEauthorblockA{\IEEEauthorrefmark{2} University of California Los Angeles, CA, USA}

\IEEEauthorblockA{\IEEEauthorrefmark{3}~Department of Public Health Sciences, University of Virginia, Charlottesville, VA, USA}

\{\href{mailto:jz4kg@virginia.edu}{jz4kg},
\href{mailto:kk7nc@virginia.edu}{kk7nc},
\href{mailto:mob3f@virginia.edu}{mob3f}, 
\href{mailto:jhh5y@virginia.edu}{jhh5y}, 
\href{mailto:jem4yb@virginia.edu}{jem4yb}, 
\href{mailto:lb3dp@virginia.edu}{lb3dp}\}@virginia.edu}

\maketitle

\begin{abstract}
Chronic kidney disease (CKD) is a gradual loss of renal function over time, and it increases the risk of mortality, decreased quality of life, as well as serious complications. The prevalence of CKD has been increasing in the last couple of decades, which is partly due to the increased prevalence of diabetes and hypertension. To accurately detect CKD in diabetic patients, we propose a novel framework to learn sparse longitudinal representations of patients' medical records. The proposed method is also compared with widely used baselines such as Aggregated Frequency Vector and Bag-of-Pattern in Sequences on real EHR data, and the experimental results indicate that the proposed model achieves higher predictive performance. Additionally, the learned representations are interpreted and visualized to bring clinical insights. 
\end{abstract}

\begin{IEEEkeywords}
Chronic Kidney Disease (CKD), Sparse Longitudinal Representations, Early Detection, CKD in Diabetic Patients, Machine Learning
\end{IEEEkeywords}
\section{Introduction}

Electronic health systems have been widely implemented in the United States and across the world \cite{jha2009use}. The availability of tremendous health data provides promising opportunities for public health research. In particular, electronic health record (EHR) data have become very popular in clinical decision support systems. They have the potential to improve healthcare by using the data and analytics for prevention, diagnosis, and treatment of diseases. However, many challenges exist when utilizing EHR data for such tasks. In this paper, we highlight the challenges of using EHR data in diseases modeling, and we explore how it can be used to improve early detection of CKD in diabetic patients. We propose a novel framework to learn a sparse longitudinal representation of patients' medical histories in EHR databases. It captures the important patterns in EHR data, thus helping improve clinical decision making and early detection of health outcomes.  
The wide implementation of electronic health systems provides a research opportunity for understanding unknown disease correlations and for better characterization of patients. However, it is challenging to analyze EHR data for a variety of reasons, presented as follows~\cite{lee2011mining, jensen2012mining}:

\textbf{Complexity:} EHR data are generally high-dimensional and sparse, and contain information collected from multiple sources, which makes it hard to integrate them into a universal feature space while preserving all useful information. In addition to single clinical events, there are multivariate and nested sequences stored in EHR systems. It is difficult for machines to understand the data's underlying patterns.
\begin{figure}
\centering
\vspace{-10pt}
\includegraphics[width=0.47\textwidth]{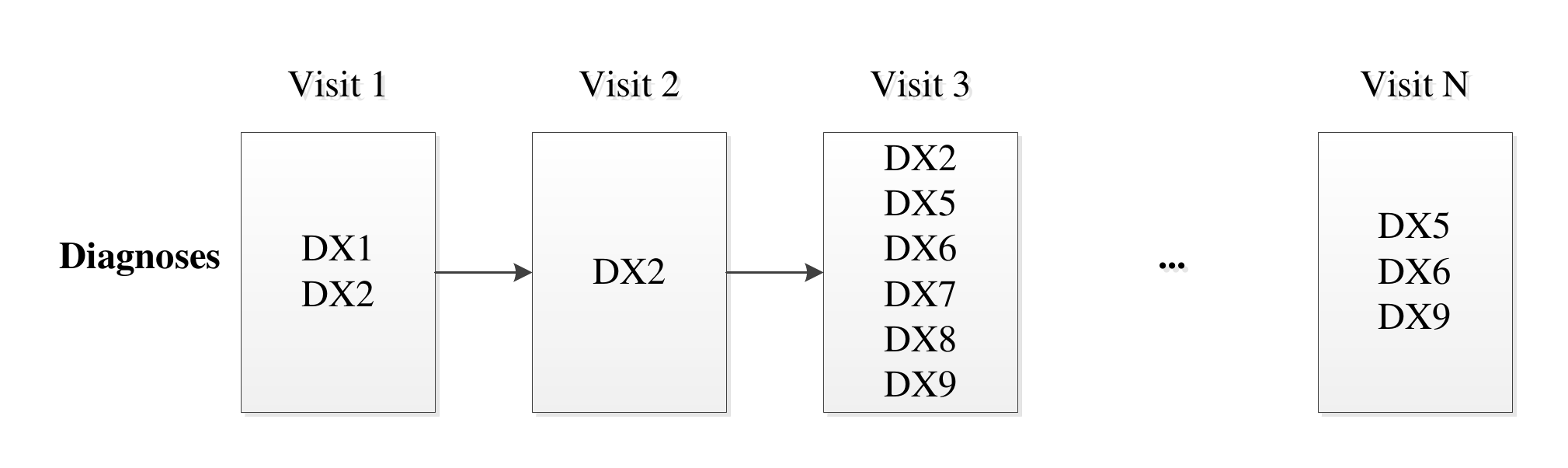}
\vspace{-10pt}
\caption{An example of a patient's profile in EHR systems}
\label{fig:3.1}
\vspace{-10pt}
\end{figure}

\textbf{Heterogeneity:} EHR systems store huge amounts of health histories of many patients, whose characteristics and medical conditions could be very diverse from each other. The heterogeneity in patients' medical histories and characteristics brings another challenge to analytical tools for clinical decision support.
Apart from the heterogeneity between patients, there is also a fairly high level of variation in clinicians' practices. For instance, if a diabetic patient visits a physician for an acute condition unrelated to the chronic disease, the physician might or might not code diabetes in the electronic record for this patient. In addition, a patient with chronic diseases might schedule a regular doctor visit every $30$ days while a similar patient might have a doctor visit every $45$ days. These heterogeneities make it challenging to mine EHR data. 

\textbf{Interpretability:} Physicians make clinical decisions based on precise knowledge of the patients' health conditions, medical histories, and other characteristics. Patterns learned by machine learning methods from EHR databases provide supplemental or even unknown knowledge to support decision making~\cite{kowsari2019text}. Thus, in addition to improving clinical decision support, it requires the patterns discovered by machines to be interpretable to humans for further validation and knowledge discovery.

\textbf{Time-invariance:} Patients' medical histories are not aligned in absolute time. Thus, representations that capture information in a relative time perspective are needed for further modeling.

\textbf{Scalability:} EHR systems store medical histories from heterogeneous data sources of large numbers of patients; patients with chronic conditions can have very complex and long histories. This aspect contributes to the large scale data contained in the EHR database, and ultimately requires efficient computational methods for analysis.

Hence, EHR data is complex which makes it a challenging task to extract useful information for further modeling and analytics. It is then important to design new feature representation frameworks to best represent the sparse EHR data. 


\section{Background}

Chronic kidney disease (CKD) is a general term of heterogeneous disorders characterizing the gradual loss of renal function over time~\cite{levey2012chronic, levey2003national, ckd_mayo}. There are five stages of CKD: Stage 1 - kidney damage with normal kidney function; Stage 2 - kidney damage with mild loss of kidney function; Stage 3 - mild to severe loss of kidney function; Stage 4 - severe loss of kidney function; and Stage 5 - also known as end stage renal disease (ESRD), which indicates kidney failure requiring dialysis or transplant for survival~\cite{ckd_nih}. CKD could eventually progress to kidney failure, which is fatal without dialysis or a kidney transplant~\cite{ckd_mayo}. It increases the risk of mortality, decreased quality of life, as well as serious complications, such as cardiovascular disease, anemia, mineral and bone disorders, fractures, and cognitive decline~\cite{jha2013chronic}. The worldwide prevalence of CKD in the general population is $13.4\%$, and it imposes a huge economic burden globally~\cite{hill2016global}. According to the Centers for Disease Control and Prevention (CDC), $30$ million Americans, i.e., $15\%$ of the adult population, are estimated to have CKD. The prevalence of CKD has been increasing in the last couple of decades, which is partly due to the increased prevalence of diabetes and hypertension~\cite{coresh2007prevalence}. Medicare spending exceeds $\$50$ billion in 2013 for CKD patients ages $65$ or older, which is $20\%$ of all medical spending for this age group~\cite{ckd_nih}. 

Diabetes and high blood pressure are the main risk factors of CKD and close to half of the CKD patients also have diabetes and/or self-reported cardiovascular disease~\cite{ckd_nih}. This study by Bailey et al. confirms the high prevalence of CKD, $43.5\%$, in patients with type $2$ diabetes~\cite{naranjo2011patients}. However, CKD starts with impaired renal function and is usually asymptomatic until the later stages~\cite{ckd_mayo, hill2016global}. According to the CDC, almost half of the patients with severely reduced kidney function but not on dialysis are not aware of having CKD, and it is also unknown to approximately $96\%$ of people with kidney damage or mildly reduced renal function~\cite{ckd_cdc2}.

Early detection and treatment of CKD can slow the progression of kidney damage by controlling the underlying cause~\cite{ckd_mayo, jha2013chronic}. Early medical interventions can also prevent the risk of complications, especially cardiovascular disease, which is the leading cause of morbidity and mortality in dialysis patients. It has been widely demonstrated that interventions in the conservative phases of CKD are more effective and should be performed as early as possible~\cite{locatelli2002importance}. 
Considering the prevalence of CKD among diabetic patients, the difficulty in recognizing it in the early stage, and the extreme importance of early interventions, we apply the proposed representation learning frameworks, \SLR{} and \WBSLR{}, to this clinically meaningful problem, i.e., recognizing misdiagnosed and undiagnosed CKD in diabetic patients. In the rest of this paper, we use the terminology ``prediction'' to indicate the early recognition of undiagnosed or misdiagnosed CKD in diabetic patients.
\begin{figure*}
\centering
\includegraphics[width=0.79\textwidth]{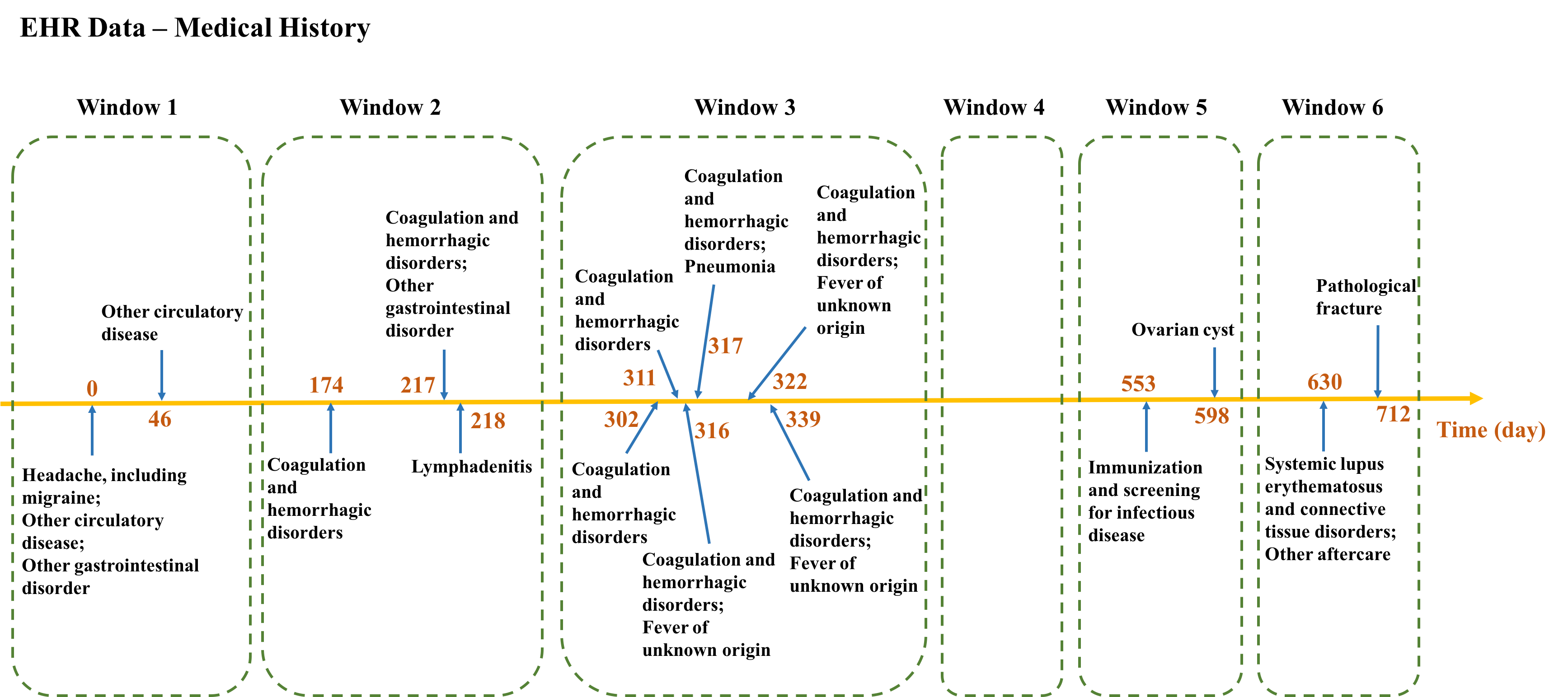}
\caption{The EHR data of an example patient and the construction of time windows}
\label{fig3.2}
\end{figure*}
\section{Methods}\label{sec:method}
In this paper, we present two representation learning frameworks namely Sparse Longitudinal Representation (SLR) and Weighted Bagging of Sparse Longitudinal Representation (WB-SLR) frameworks aiming both to find a unified representation of EHR data for an accurate prediction. The problem formulation and the framework design of SLR and WB-SLR will be presented below. 
\subsection{SLR: Sparse Longitudinal Representation}\label{sec:SLR}
Given (1) the training set, a patient's health record, consisting of a sequence of visits with each visit containing a set of medical codes (shown in Figure~\ref{fig:3.1}), namely, $s_i$, for patient $i$, and (2) each patient's label (i.e., $1$ or $0$) representing the outcome (e.g., $y_i=1$ indicating that patient $i$ has a diagnosis of disease of interest or the targeted outcome in their medical history), our research problem is to find a method to transform each $s_i$ to a unified representation $x_i$, where $x_i \in R^{d}$, so as to maximize the accuracy of any arbitrary classifier.

\begin{equation}
\label{eq3.0}
    \mathrm{min}  \displaystyle\sum_{i=1}^N|y_i - \mathit{classifier}(x_i)|,
\end{equation}
where $N$ is the number of patients and the function $\mathit{classifier}: R^{d}\to\{0, 1\}$ refers to an arbitrary binary classifier on top of feature space $R^{d}$.

\begin{figure}
\centering
\includegraphics[width=0.47\textwidth]{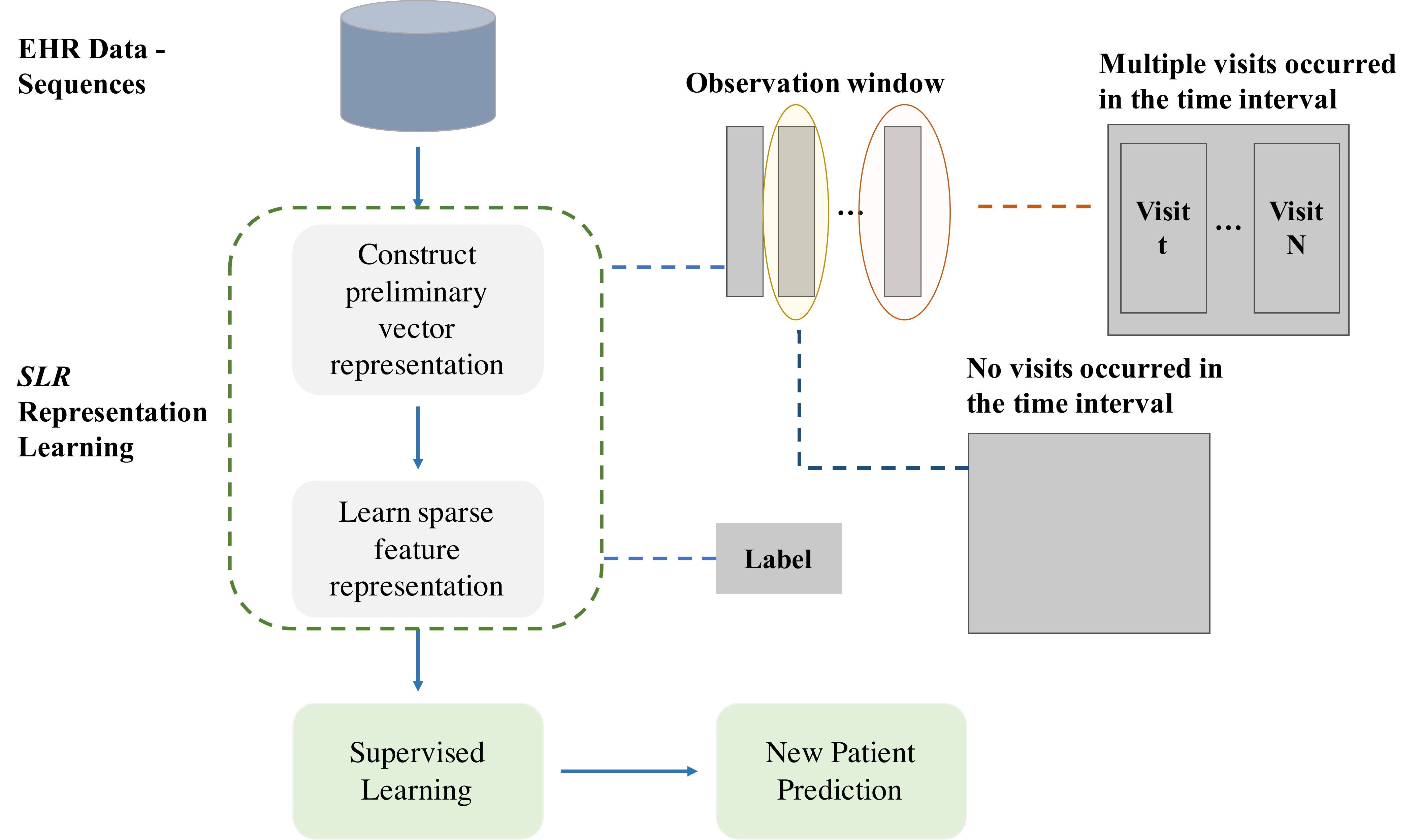}
\caption{The SLR representation learning framework}\label{fig3_system}

\end{figure}

The \SLR{} framework contains the following three components illustrated in Figure~\ref{fig3_system}.

\textbf{Representation Learning:} Given each patient's medical history and the corresponding label, this step transforms each patient's record into a unified vector space. Further, a sparse group lasso based algorithm~\cite{simon2013sparse} is employed to learn a sparse longitudinal representation of the patient's medical history.

\textbf{Supervised Learning:} Given the learned data representation from Step 1 and the label of each patient, this step trains a predictive model using supervised learning algorithms, such as logistic regression and random forest.

\textbf{New Patient Prediction:} Given a feature representation from Step 1, this step uses the predictive model (from Step 2) to predict the new patient's label. The outcome of this step is $1$ or $0$, which refers to whether the patient will have the targeted health outcome. 

\subsubsection{Preliminary Vector Representation}\label{sec:SLR_PVP}~\\
As presented in Figure~\ref{fig:3.1}, the medical history of patient $i$ is a sequence $s_i$ with multiple visits and there are one or more clinical events during each visit. Commonly, the time intervals between visits are different within and across patients. One straightforward way of representing this data is to use the counts or binary indicators of the clinical events in the sequence. However, temporal information is omitted in this approach. To cope with this issue, we develop an itemset representation which truncates the sequence $s_i$ into $T$ time windows with even intervals and an itemset of clinical events is constructed for each time window $t_j$. Then, the counts of clinical events in each time window is computed and concatenated as a preliminary representation $x_i$ of patient $i$'s medical history, i.e., $x_i = \{v_{i1}, v_{i2}, \cdots, v_{ij}, \cdots, v_{iT}\}$ and the associated time windows are $\{t_1, t_2, \cdots, t_j, \cdots, t_T\}$. Here, $v_{ij} = \{c_{j1}^{(i)}, c_{j2}^{(i)}, \cdots, c_{jp}^{(i)}, \cdots, c_{jP}^{(i)}\}$ is the count vector of clinical events $A_1, A_2, \cdots, A_p, \cdots, A_P$ at time window $t_j$, where $P$ is the total number of distinct clinical events. Thus, all elements in a patient's preliminary representation are presented in Table~\ref{tab3.1}.

\begin{table}[ht]
\centering
\caption{The preliminary vector representation $x_i$ of Patient~$i$}
\label{tab3.1}
\begin{tabular}{llllllll}
       & $A_1$  & $A_2$  &$A_3$ & $\cdots$  &$A_p$ &$\cdots$  &$A_P$\\ \cline{2-8}
{$t_1$} & $c_{11}^{(i)}$ & $c_{12}^{(i)}$ & $c_{13}^{(i)}$ &$\cdots$ & $c_{1p}^{(i)}$& $\cdots$ & $c_{1P}^{(i)}$\\ 										
{$t_2$} & $c_{21}^{(i)}$ & $c_{22}^{(i)}$ & $c_{23}^{(i)}$ &$\cdots$ & $c_{2p}^{(i)}$& $\cdots$ & $c_{2P}^{(i)}$\\ 
{$t_3$} & $c_{31}^{(i)}$ & $c_{32}^{(i)}$ & $c_{33}^{(i)}$ &$\cdots$ & $c_{3p}^{(i)}$& $\cdots$ & $c_{3P}^{(i)}$\\ 
{$\cdots$} & $\cdots$ & $\cdots$ & $\cdots$ &$\cdots$ & $\cdots$& $\cdots$ & $\cdots$\\ 
{$t_j$} & $c_{j1}^{(i)}$ & $c_{j2}^{(i)}$ & $c_{j3}^{(i)}$ &$\cdots$ & $c_{jp}^{(i)}$& $\cdots$ & $c_{jP}^{(i)}$\\ 
{$\cdots$} & $\cdots$ & $\cdots$ & $\cdots$ &$\cdots$ & $\cdots$& $\cdots$ & $\cdots$\\  
{$t_T$} & $c_{T1}^{(i)}$ & $c_{T1}^{(i)}$ & $c_{T1}^{(i)}$ &$\cdots$ & $c_{Tp}^{(i)}$& $\cdots$ & $c_{TP}^{(i)}$\\ \hline
\end{tabular}
\end{table}

Figure~\ref{fig3.2} illustrates the medical history of an example patient and the construction of itemsets with the associated time windows. In this example, the size of each time window is $4$ months and an itemset is constructed for each time window. Then, the counts of clinical events in the itemsets are computed to represent the medical record at each window, while the events not present in the time window are assigned a value $0$. Finally, the count vectors of clinical events in each of the time windows are concatenated to construct a preliminary representation of the patient's medical history.

\subsubsection{SLR Representation Learning Algorithm}~\\
With the preliminary data representation, $x_i = \{v_{i1}, \cdots, v_{iT}\}$, the proposed method based on the sparse group lasso algorithm~\cite{simon2013sparse} aims at learning a sparse set of events in the longitudinal order, as presented in Equation~\ref{eq3.1}.

\begin{equation}
\label{eq3.1}
min_\omega L(X\omega, y) + (1-\alpha) \lambda \displaystyle\sum_{j=1}^T \sqrt[]{P}||\omega_{j}||_2 + \alpha \lambda ||\omega||_1
\end{equation}
where $P$ is the number of distinct clinical events. $j$ is the $j$th time window ($j \in \{1, 2, \cdots, T\}$) and $\omega_{j}=\{\omega_{j1}, \omega_{j2}, \cdots, \omega_{jP}\}$. $L(X\omega, y)$ is the loss function shown in Equation~\ref{eq3.2}.

\begin{equation}
\label{eq3.2}
L(X\omega, y) = \frac{1}{N}\displaystyle\sum_{i=1}^N log(1+exp(-y_ix_i^\intercal \omega))
\end{equation}
where $N$ is the total number of patients. The objective function in \SLR{} learning consists of three parts: error minimization, $l2$ penalty on each time window of features, and a $l1$ sparsity term on each feature throughout the entire medical history. Thus, $SLR$ learns a sparse longitudinal representation of the medical history by minimizing the error in Equation~\ref{eq3.1}.  We propose Algorithm~\ref{alg:1} to learn a sparse longitudinal representation from the EHR data. 

\begin{algorithm}[!htb]
\SetKwInOut{Input}{Input}\SetKwInOut{Output}{Output}
\Input{Set of sequences $S$, set of events $A$, number of time intervals $T$, number of patients $N$, number of distinct events $P$}
\Output{Set of selected events $A_s$}
\Begin{
 {\bf Construct Preliminary Data Representation $X$:}\\
 {Initialization:}\\
 $X \leftarrow \phi$;\\
 \For { $i = 1 $ to $N$} {
 $x_i \leftarrow \phi$\\
 	 \For { $j = 1 $ to $T$} {
     $c_j^{(i)} \leftarrow \{c_{j1}^{(i)}, \cdots, c_{jp}^{(i)}, \cdots, c_{jP}^{(i)}\}$ \\
     $x_i \leftarrow \{x_i, c_j^{(i)} \}$\\
     } 
      $X_i \leftarrow x_i$\\
 }
  {\bf Learn \SLR{} Representation $A_s$:}\\
  {Initialization:}\\
  $A_s \leftarrow \phi$;\\
  {Block coordinate descent to optimize~(\ref{eq3.1}): $\hat{\omega}$}\\
 	 \For {$j = 1 $ to $T$} {
     $\omega_{j} \leftarrow \{\omega_{j1}, \cdots, \omega_{jp}, \cdots, \omega_{jP}\}$ \\
     \If {$\omega_{j} \neq \mathbf{0}$} {
      \For {$p = 1$ to $P$} {
       \If {$\omega_{jp} \neq 0$} {
           $A_s \leftarrow A_s \cup \{A_{jp}\}$ 
       }
     }
     }
     }
\Return{$A_s$}}
\caption{\SLR{} Representation Learning}
\label{alg:1}
\end{algorithm}


\begin{figure}
\centering
\includegraphics[width=0.47\textwidth]{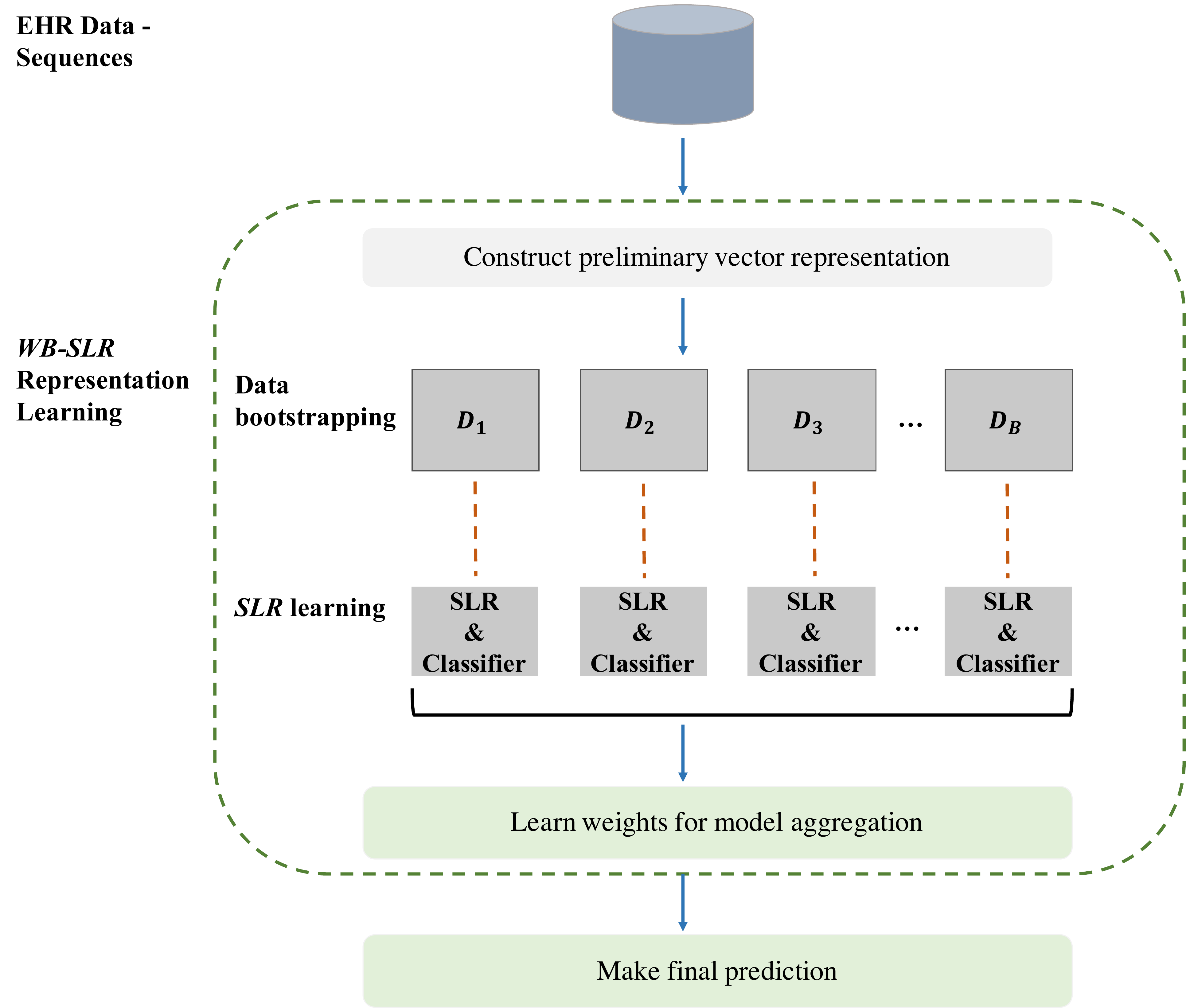}
\caption{The \WBSLR{} representation learning framework}
\label{fig4_system}
\end{figure}

This problem is the sum of a convex function and a penalty term, so that the coordinate descent algorithm is employed to learn the global optimum~\cite{simon2013sparse}. When $\alpha=1$, Equation~\ref{eq3.1} becomes lasso, which learns a sparse representation of the clinical events, while it becomes group lasso when $\alpha=0$, which is equivalent to ridge regression on groups of features. Here, $\alpha$ and $\lambda$ are two tuning parameters to control the degree of sparsity in the \SLR{} representation. 

\subsection{WB-SLR:  Weighted Bagging of Sparse Longitudinal Representation}\label{sec:WBSLR}

Given patients' longitudinal EHR data, the research problem remains the same as in Section~\ref{sec:SLR}, i.e., to find a method to transform each sequence $s_i$ into a unified representation for an accurate prediction. In this section, we propose \WBSLR{} which learns a weighted bagging of \SLR{}s to provide a more stable and comprehensive representation.

The \WBSLR{} framework consists of four steps and Figure~\ref{fig4_system} presents a graphical illustration of the designed framework.

\textbf{Constructing Preliminary Representation} --- Given each patient's medical record, this step transforms a sequence into the preliminary vector representation of clinical events at multiple time windows, as elaborated in~\ref{sec:SLR_PVP}.

\textbf{Learning \SLR{} Ensemble} --- Given the training set of preliminary vector representations, this step first performs bootstrapping on the training set. Then, a \SLR{} representation and an associated ensemble classifiers are learned on each bootstrapped sample such as any kind of the ensemble learning~\cite{henriksson2015modeling,kowsari2018rmdl,heidarysafa2018improvement}. 

\textbf{Learning Weights for Model Aggregation} --- Given the ensemble of classifiers and associated \SLR{}s, the classifiers are aggregated linearly with weights such that the oob error is minimized.

\textbf{Making Final Prediction} --- This step makes the final prediction for each patient by weighted aggregation of the output from each single classifier in the ensemble.

\subsubsection{\WBSLR{} Representation Learning Algorithm}
\label{sec4.2}~\\
In this section, we elaborate the details of the \WBSLR{} representation learning framework consisting of three parts -- \SLR{} ensemble learning, weighted model selection, and new patient prediction, as described in the following.

\textbf{\SLR{} Ensemble Learning} 
--- There are two steps to obtain the ensemble of \SLR{}s and the associated classifiers:

\subsubsection*{Constructing data bootstraps} Given the training set of preliminary vector representation $D$, we sample $B$ bootstraps, denoted as 
$\{D_1, \cdots, D_b, \cdots, D_B\}$. Each bootstrap $D_b$ is created by randomly sampling with replacement from $D$.

\subsubsection*{Learning \SLR{}s} Given bootstrap $D_b$, a sparse representation $l_b$ is learned using the \SLR{} framework proposed in Section~\ref{sec:SLR}. Additionally, an associated classifier $C(l_b, D_b)$ is trained to make prediction according to Equation~\ref{eq3.1}.

Accordingly, a classifier and the associated \SLR{} representation are learned on each bootstrap, which are the basis for weighted model selection. 

\textbf{Weighted Model Aggregation} 
--- In this part, we provide the details of model aggregation using oob weighting, which is derived from the work by Rao and Tibshirani~\cite{rao1997out}. The steps are elaborated as follows:

\begin{figure}
\centering
\includegraphics[width=0.47\textwidth]{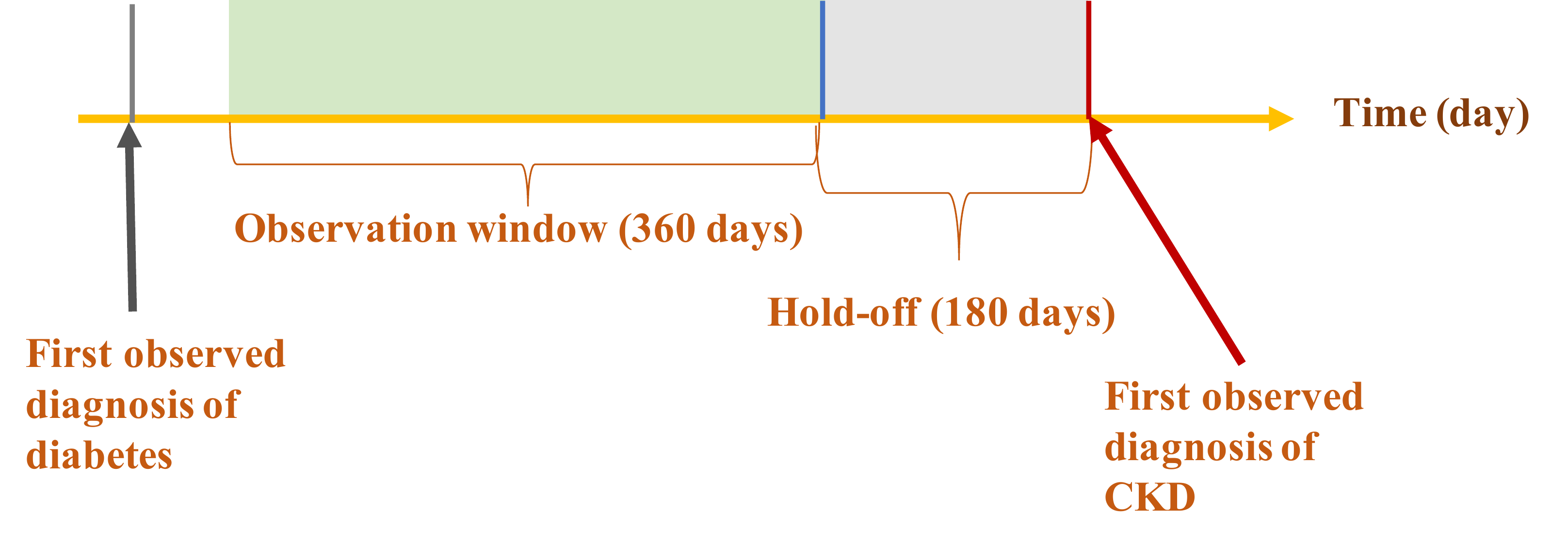}
\caption{A graphical illustration of the experimental setting for the comorbid risk prediction of CKD in diabetic patients}
\label{fig4.setting}
\end{figure}

\subsubsection*{Prediction on oob samples} Given the bootstraps $\{D_1, \cdots,$ $D_b, \cdots, D_B\}$ and the training set $D$, we use $K_i$ to denote the indices of the bootstrapped samples that do not contain patient $i$. Then, each classifier trained based on the bootstrapped samples in $K_i$ makes a prediction on the label of observation $i$. The final prediction on the oob sample is aggregated with the following equation:

\begin{equation}
\label{eq4.1}
\hat{y}_i (W) = \frac{1}{|K_i|}\displaystyle\sum_{b \in K_i}\omega_b C(l_b, D_b)
\end{equation}
where $W=\{\omega_1, \omega_2, \cdots, \omega_b, \cdots, \omega_B\}$.

\begin{table*}[!t]
\centering
\caption{The predictive performance of baselines and the proposed \SLR{} and \WBSLR{} frameworks}
\label{tab4_result}
\begin{tabular}{cccccc}
\hline
                                      &                  & Sensitivity                 & Specificity                  & AUC  & F2 score\\ \cline{3-6}
\multirow{2}{*}{Aggregated Frequency Vector~(AFV)}                  & LR               & $0.649 \pm 0.030$           &  $0.734 \pm 0.025$           & $0.712 \pm 0.025$   &  $0.660 \pm 0.037$ \\
                                      & RF  & $0.557 \pm 0.033 $          &   $0.890 \pm 0.017$          & $0.774 \pm 0.020$   & $0.671 \pm 0.026$ \\
                                      & GBT  & $0.623 \pm 0.029 $          &   $0.891 \pm 0.028$          & $0.796 \pm 0.021$   & $0.667 \pm 0.023$ \\\hline
                                      
\multirow{2}{*}{Bag-of-Pattern in Sequences~(BPS)}                  & LR               & $0.592 \pm 0.042$           & $0.728 \pm 0.026$            & $0.757 \pm 0.022$    & $0.592 \pm 0.031$ \\
                                      & RF & $\mathbf{0.975 \pm 0.011}$           &  $0.287 \pm 0.026$           & $0.732 \pm 0.027$   &   $0.803 \pm 0.020$ \\
                                      & GBT  & $0.831 \pm 0.026 $          &   $0.684 \pm 0.026$          & $0.807 \pm 0.018$   & $0.749 \pm 0.024$ \\\hline

\multirow{2}{*}{Aggregated Transition Vector
~(ATV)}                  & LR               & $0.500 \pm 0.033$            &  $0.888 \pm 0.027$          & $0.697 \pm 0.026$   &  $0.546 \pm 0.035$  \\
                                      & RF & $0.598 \pm 0.032$            & $0.842 \pm 0.022$           & $0.791 \pm 0.021$  &  $0.712 \pm 0.028$  \\ & 
                                      GBT  & $0.558 \pm 0.031 $          &   $0.885 \pm 0.017$          & $0.796 \pm 0.025$   & $0.722 \pm 0.024$ \\\hline
                                      
\multirow{2}{*}{Sparse Longitudinal Representations~(\SLR{})}                  & LR               & $0.747 \pm 0.035$            &  $0.899 \pm 0.018$          & $0.835 \pm 0.015$  &  $0.772 \pm 0.031$  \\
                                      & RF & $0.753 \pm 0.033$            &  $\mathbf{0.903 \pm 0.019}$          & $0.847 \pm 0.026$  & $0.799 \pm 0.030$    \\ 
                                      & GBT  & $0.857 \pm 0.024 $          &   $0.775 \pm 0.025$          & $0.877 \pm 0.017$   & $0.810 \pm 0.016$ \\\hline
                                     
\multicolumn{2}{c}{Bagged \SLR{}}              & $0.829 \pm 0.023$            &  $0.865 \pm 0.020$          & $0.842 \pm 0.020$  &   $0.818 \pm 0.028$\\ \hline
                                      
\multicolumn{2}{c}{\WBSLR{}}                & $0.835 \pm 0.025$            & $0.852 \pm 0.012$           & $\mathbf{0.891 \pm 0.018}$ &  $\mathbf{0.820 \pm 0.027}$\\ \hline 

\end{tabular}

\end{table*}

\subsubsection*{Learning of oob weighting} --- Given $(\hat{y}_i(W), y_i)$ for observation $i$ in the training set, we optimize the weights $W$ of all classifiers in the ensemble.
Considering that this is a classification problem, we employ the negative log-likelihood as the objective function with the constraints that $\omega_b \geq 0$ for all $b \in \{1, 2, \cdots, B\}$, as presented in Equation~\ref{eq4.3}.
\begin{equation}
\label{eq4.3}
\begin{aligned}
& \underset{W}{\text{minimize}}
& & -\displaystyle\sum_{i=1}^N y_i log\hat{y}_i(W) + (1-y_i) log(1-\hat{y}_i (W)) \\
& \text{subject to}
& & W \geq \mathbf{0}.
\end{aligned}
\end{equation}
Then, the truncated Newton method, Newton conjugate gradient algorithm~\cite{shewchuk1994introduction}, is utilized to solve the optimization problem in Equation~\ref{eq4.3}, and the optimal weights are denoted as $\hat{W} = \{\hat{\omega}_1, \cdots, \hat{\omega}_b, \cdots, \hat{\omega}_B\}$.

\textbf{New Patient Prediction} 
--- Given the learned representations and classifiers in the ensemble with associated weights, we make final prediction for a new patient $i$ as presented in Equation~\ref{eq4.4}. 
\begin{equation}
\label{eq4.4}
\hat{y}_i (\hat{W}) = \frac{1}{B}\displaystyle\sum_{b=1}^B\hat{\omega}_b C(l_b, D_b)
\end{equation}

Thus, the \WBSLR{} framework introduces a weighted bagging of the \SLR{}s to obtain the final output in order to achieve more stable and more accurate prediction.

\section{Evaluation}

\subsection{Experimental Design}
\label{sec3.3.2}
In this work, we use the de-identified EHR data of $75$ months beginning in September $2010$ from the University of Virginia Health System. This dataset contains $2{,}343{,}651$ inpatient and outpatient visits of $473{,}915$ distinct patients. We focus our experiment on the EHR's diagnosis codes which are coded primarily in ICD-$9$ and a small portion are ICD-$10$ codes. The raw diagnoses codes are further clustered into $283$ groups according to the Clinical Classification Software from the Agency for Healthcare Research and Quality (AHRQ)~\cite{ahrq}. 

In the experiment, we define the observation window and prediction period to validate the proposed method. We first extract all patients with diabetes diagnoses in their medical record. For diabetic patients with CKD diagnoses to be included in the cohort, their diabetes condition has to be diagnosed at least $1.5$ years prior to the first observed diagnoses of CKD. Thus, it allows adequate time to observe the health conditions to construct features and allows the hold-off window to be used for the purpose of early detection. 

For positive patients, their visits during $1.5$ to $0.5$ year prior to the first observed CKD are used to construct features for prediction. The visits within $6$ months of the first observed CKD diagnosis are excluded. For diabetic patients without observed CKD diagnoses, we exclude the individuals when their visit histories are shorter than $2.5$ years after the first observed diagnosis of diabetes. Visits occurred in the first $12$ months are used as observations for modeling, and the next $0.5$ year is the hold-off window same as that for the positive patients. The following $1$ year is monitored to ensure the patients are not diagnosed with CKD in the near future. Thus, a $12$-month observation window is constructed for patients in each group, and a prediction window is defined to allow observation of the outcome. To better illustrate the experimental setting, we present the observation window, hold-off and onset of outcome event in Figure~\ref{fig4.setting}.

To validate the proposed representation learning frameworks, we compare the prediction performance of the proposed models with 4 baseline approaches as follows.
\begin{itemize}
\item Aggregated Frequency Vector (AFV): A patient's record is represented as a count vector of medical events in the observation window. Each dimension is associated with a distinct medical event in patients' EHR data.
\item Bag-of-pattern in Sequences (BPS): This method runs a widely used sequence pattern mining algorithm, CM-SPADE~\cite{fournier2014fast}, to discover all frequent patterns in patients' EHR data. Thus, a patient's record is represented with a frequency vector of the discovered frequent patterns.
\item Aggregated Transition Vector (ATV): The pairwise transitions between medical event pairs are counted and a patient's record is represented with a frequency vector of pairwise transitions. 
\item Bagged \SLR{}: It first learns a \SLR{} representation and an associated classifier on each bootstrap sample. Then, the final prediction is obtained by majority voting over all the classifiers.
\end{itemize}

With the proposed \SLR{}, \WBSLR{}, and baseline representations of patients, we model the comorbid risk of CKD using three classifiers, logistic regression with $l1$ penalty (LR), random forest (RF), and gradient boosting trees (GBT) with 20 trees, respectively. The hyperparameters of classifiers are tuned on the validation set. The performance of baselines and the proposed frameworks on the comorbid risk prediction task is evaluated with AUC (area under curve), sensitivity, specificity, and F2 score.The
F2 measure assigns sensitivity twice as much weight as Positive Predictive Value~(PPV) and can be interpreted as a measure that is biased toward sensitivity~\cite{lu2008ontology}. Each experiment is repeated $50$ times and we calculate the averages and standard deviations of the above metrics, respectively.



\subsection{Experimental Results}
The predictive performance of the classifiers based on the baseline representations and the proposed \SLR{} and \WBSLR{} frameworks are presented in Table~\ref{tab4_result}. Here, the results shown are based on the \SLR{} and \WBSLR{} learned with $\alpha=0.7$ and $\lambda=0.0005$. 
According to the results table, the predictive performance of classifiers based on \WBSLR{} outperform the baselines in terms of AUC. The classifier based on \WBSLR{} achieves a relatively higher AUC by approximately $5\%$ than the bagged \SLR{}. The \WBSLR{} achieves a highly balanced prediction result on the target and control groups by comparing the sensitivity and specificity. With the baseline representations, the GBT and RF classifiers generally outperform LR classifiers in terms of AUC and F2 score. In general, the GBT models demonstrates a slightly better performance than RFs. The performance between the first four baselines are not significantly different, while the \SLR{} based representations are capable of improving the prediction performance significantly. In terms of AUC, the bagged \SLR{} achieves a more accurate prediction than the LR classifier based on \SLR{}, while the RF and GBT models based on \SLR{} outperform the bagged \SLR{}. This finding is consistent with the properties of bagging and random forest. 

In general, the proposed \WBSLR{} is able to achieve a more accurate prediction compared to baselines and we hypothesize that the improvement could be more significant if there are more observations to allow better weight learning for model aggregation.



\begin{figure*}[!t]
\centering
\includegraphics[width=0.69\textwidth]{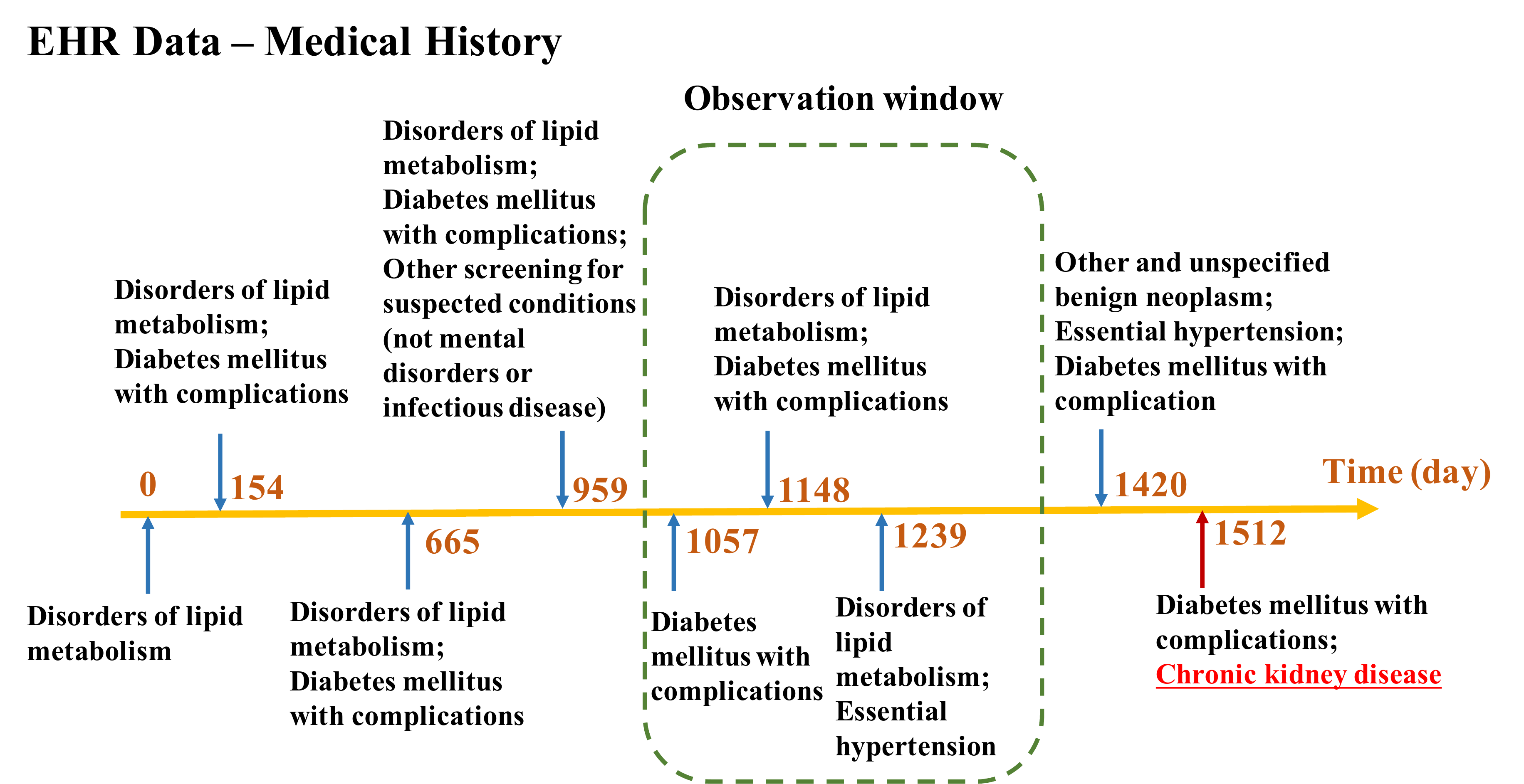}
\caption{The medical history of an example diabetic patient who developed CKD later}
\vspace{-10pt}
\label{fig4_example1}
\end{figure*}

\subsection{Visualization \& Interpretation}
\label{sec:Visualization}

In addition to predictive performance, we elaborate the selected features by the \SLR{} with highest weight in \WBSLR{} in the early detection of CKD among diabetic patients. In general, we observe that the diagnoses temporally closer to the observation of outcome are more positively correlated with it while the selected features that are negatively correlated are mostly in the first time window, i.e., occurred much earlier than the outcome onset. Negatively correlated events in the very early time window include \textit{non-specific chest pain}, \textit{osteoarthritis}, \textit{thyroid disorders}, \textit{other connective tissue disease}, \textit{essential hypertension}, \textit{diabetes with/without complications}, and so forth.

However, the diabetic diagnoses become a very important positive risk factor when it is later in the time window, especially the \textit{diabetes mellitus with complications}. Other positively correlated risk factors throughout the observation window are \textit{deficiency and other anemia}, \textit{other aftercare}, \textit{other injuries and conditions due to external causes}, and heart and cardiovascular diseases, including \textit{congestive heart failure}, \textit{cardiac dysrhythmias}, \textit{other and ill-defined heart disease}, etc. Here, \textit{deficiency and other anemia} refers to iron deficiency anemias, other deficiency anemias, and/or hereditary anemias according to ~\cite{ahrq}, which are found to be correlated to CKD according to~\cite{babitt2012mechanisms, gotloib2006iron, mcclellan2004prevalence}. In fact, previous literature have shown that the prevalence of anemia increases as kidney function decreases ~\cite{mcclellan2004prevalence, hsu2002epidemiology, patel2010vitamin}. 

Additionally, we observe that \textit{essential hypertension} and a group of cardiovascular and heart diseases are strongly correlated with the onset of CKD. Previous research have demonstrated that \textit{essential hypertension} is one of the leading causes of CKD together with diabetes~\cite{atkins2005epidemiology}. According to~\cite{liu2014cardiovascular}, there is a close interrelation between cardiovascular disease and kidney disease, and the disease of one organ may cause dysfunction of the other, which could ultimately lead to the failure of both organs~\cite{liu2014cardiovascular}. Additionally, both cardiovascular and heart diseases and kidney disease are complications of diabetes. It is also likely that the diabetic patients developed cardiovascular and heart problems because of uncontrolled diabetes such that we observe the diagnoses of CKD later as another complication of diabetes, apart from the damages to renal function by hypertension and cardiovascular diseases. 

\textit{Other nervous system disorders}, \textit{other aftercare} and \textit{other injuries and conditions due to external causes} are also shown to be positively correlated with CKD. According to ~\cite{krishnan2009neurological, arnold2016neurological}, neurological complications are prevalent in CKD patients and occur in almost all patients with severe CKD, which might affect both the central and peripheral nervous systems. Here, the \textit{other aftercare} diagnosis group includes aftercare following surgeries and long-term (current) use of drugs, such as insulin. In the diagnosis group \textit{other injuries and conditions due to external causes}, there are diagnoses of injuries due to accidents as well as other nonspecific abnormal toxicological findings which includes abnormal levels of drugs or heavy metals in blood, urine, or other tissue~\cite{ahrq}. Medical research has shown that exposure to heavy metals and chronic use of drugs known to be potentially nephrotoxic can lead to CKD~\cite{national2008chronic, kazanciouglu2013risk}. It is clinically counterfactal to observe \textit{disorders of lipid metabolism}, \textit{essential hypertension}, and \textit{diabetes with/without complications} being negatively correlated factors earlier in the observation window, while they become strong positive risk factors later. The potential rationale is that these diagnoses are prevalent in the early observation window in both the negative and positive patient cohorts, however, the factors making real differences are the occurrences of them later in the observation period or constant occurrences of these diagnoses. Thus, the early observation of these factors are learned as being negatively correlated to the onset of CKD.

To provide a more straightforward understanding of the learned representation and its benefit to the comorbid risk prediction of CKD in diabetic patients, we present the medical history of an example diabetic patients in Figure~\ref{fig4_example1} which shows the EHR data of a diabetic patient who developed CKD later. We observe that \textit{disorders of lipid metabolism} and \textit{diabetes mellitus with complications} occur repeatedly in this patient's medical record, and those are found to be positively correlated with comorbid CKD. The example patient is predicted correctly using the proposed \WBSLR{} representation learning framework. 

In general, the learned \WBSLR{} representation is consistent with the previous medical research. In addition to improving the accuracy of the early detection of CKD in diabetic patients, this research could also be used to deepen clinical understanding of disease correlations.

\section{Conclusion}

In this paper, we propose novel representation learning frameworks, \SLR{} and \WBSLR{}, to learn comprehensive and stable representations of patients' EHR data. \SLR{} focuses on learning sparse representations, while \WBSLR{} further utilizes the bagging strategy and model aggregation based on oob weighting as an improvement of \SLR{}. We apply these frameworks to the early detection of CKD in diabetic patients using longitudinal EHR data. The experimental results demonstrate that the proposed \WBSLR{} representation learning framework is capable of achieving more accurate prediction and they also uncover the correlations between diseases which are found to be consistent with previous medical research.

Alternative model selection and combination methods will be explored to aggregate the ensemble of classifiers and representations learned from bootstrapped samples. Moreover, future improvement on the proposed model could employ the strategy of random forest which samples a subset of features when growing a tree. This could potentially reduce the correlation between the single models in the ensemble and further improve the prediction accuracy. Regularization is introduced to reduce multicollinearity and other potential approaches to address this issue will be explored as future directions. Again, future work will consider more specific diagnosis codes and other grouping strategies to reduce the information loss introduced by using the current AHRQ clinical classification scheme. Other types of clinical events, such as procedures and medications, could be added to learn a more comprehensive representation of patients' medical histories. In~\cite{zhang2018patient2vec}, we proposed~\textit{Patient2Vec} to learn personalized interpretable deep representations of EHRs to address more complicated data. In future work, we will explore its application to the early detection of CKD in diabetic patients.

In the experiment, we model the comorbid risk of CKD in diabetic patients with a small population under the current experimental setting, i.e., $395$ individuals are identified under the setting of having their first CKD diagnoses at least $18$ months after the first observed diabetes diagnoses. To increase the sample size, we could potentially employ the medical records of CKD patients without diabetes or with concurrent diabetes. CKD is similar in patients with or without diabetes and would most likely share some common risk factors or symptoms. Hence, there is great potential to improve the prediction performance by including those patients into the originally identified positive patients. We will also explore transfer learning approaches to transfer the knowledge learned from a similar population to address the target problem.

\section*{Acknowledgment}
This research was supported by a Jeffress Trust Award in Interdisciplinary Science.

\bibliographystyle{IEEEtran}
\bibliography{ref.bib}

\end{document}